\documentclass[letterpaper]{article} 
\usepackage{aaai2026}  
\usepackage{times}  
\usepackage{helvet}  
\usepackage{courier}  
\usepackage[hyphens]{url}  
\usepackage{graphicx} 
\usepackage{natbib}  
\usepackage{caption} 
\usepackage{amsmath}
\usepackage{algorithm}
\usepackage{algorithmic}
\usepackage{booktabs}
\usepackage{multirow}
\usepackage{newfloat}
\usepackage{listings}
\DeclareCaptionStyle{ruled}{labelfont=normalfont,labelsep=colon,strut=off} 
\floatstyle{ruled}
\newfloat{listing}{tb}{lst}{}
\floatname{listing}{Listing}
\usepackage{booktabs}
\title{\textit{SpecDiff}: Accelerating Diffusion Model Inference with Self-Speculation}
\author{
    Jiayi Pan\textsuperscript{\rm 1,2\equalcontrib},
    Jiaming Xu\textsuperscript{\rm 1,3\equalcontrib$\dagger$},
    Yongkang Zhou\textsuperscript{\rm 1,3},
    Guohao Dai\textsuperscript{\rm 1,2,3$\dagger$}
}

\affiliations{
    \textsuperscript{\rm 1}Shanghai Jiao Tong University,\\
    \textsuperscript{\rm 2}Infinigence-AI,\\
    \textsuperscript{\rm 3}Shanghai Innovation Institute\\
    \textsuperscript{$\dagger$}Corresponding author: jiamingxu@sjtu.edu.cn,  daiguohao@sjtu.edu.cn\\

}

\usepackage{bibentry}

\begin{document}

\maketitle

\begin{abstract}
Feature caching has recently emerged as a promising method for diffusion model acceleration. It effectively alleviates the inefficiency problem caused by high computational requirements by caching similar features in the inference process of the diffusion model. In this paper, we analyze existing feature caching methods from the perspective of information utilization, and point out that relying solely on historical information will lead to constrained accuracy and speed performance. And we propose a novel paradigm that introduces future information via self-speculation based on the information similarity at the same time step across different iteration times. Based on this paradigm, we present \textit{SpecDiff}, a training-free multi-level feature caching strategy including a cached feature selection algorithm and a multi-level feature classification algorithm. (1) Feature selection algorithm based on self-speculative information. \textit{SpecDiff} determines a dynamic importance score for each token based on self-speculative information and historical information, and performs cached feature selection through the importance score. (2) Multi-level feature classification algorithm based on feature importance scores. \textit{SpecDiff} classifies tokens by leveraging the differences in feature importance scores and introduces a multi-level feature calculation strategy.
Extensive experiments show that \textit{SpecDiff} achieves average 2.80×, 2.74×, and 3.17× speedup with negligible quality loss in Stable Diffusion 3, 3.5, and FLUX compared to RFlow on NVIDIA A800-80GB GPU. By merging speculative and historical information, \textit{SpecDiff} overcomes the speedup-accuracy trade-off bottleneck, pushing the Pareto frontier of speedup and accuracy in the efficient diffusion model inference.
\end{abstract}


\section{Introduction}
\label{sec:intro}

\begin{figure}[h]
    \centering
    \includegraphics[width=0.49\textwidth]{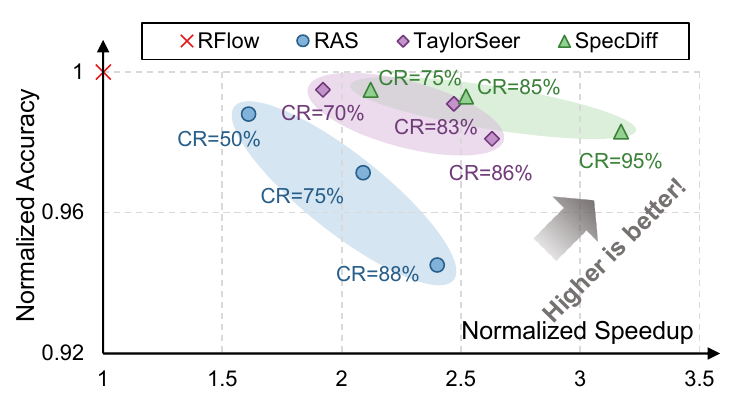} 
    \caption{Pareto frontier of accuracy and speedup towards DiT feature caching. The detailed normalized accuracy and speedup are obtained with Stable Diffusion 3 and FLUX on an NVIDIA A800-80GB GPU. CR represents the caching ratio in the configuration of feature caching methods.}
    \label{fig:pareto}
\end{figure}

Towards the advancement of multimodal artificial intelligence, the diffusion model is a typical neural network, achieving remarkable success across various domains (\textit{e.g.}, text-to-image~\cite{zhang2023text} and text-to-video~\cite{xing2024survey} generation), and significantly enabling the rapid development of numerous downstream tasks (\textit{e.g.}, content creation~\cite{wang2024diffusion}). The inference of the diffusion model is a process of continuously denoising images by iteratively executing the complete model. Each execution is a modification and refinement of the image. Driven by the scaling law~\cite{henighan2020scaling,liang2024scaling}, the diffusion model with an increasing number of parameters has demonstrated outstanding performance in many scenarios. However, this further incurs significant memory requirements and inference latency. Moreover, the input matrix to the diffusion model inference is transformed from the noise image, which has comparable row values to the model weight, leading to the compute-bound matrix multiplication operator. Reducing the actual computation workload is the direct and effective optimization method for inference acceleration.
\begin{figure*}[t] 
    \centering 
    \includegraphics[width=1\textwidth]
    {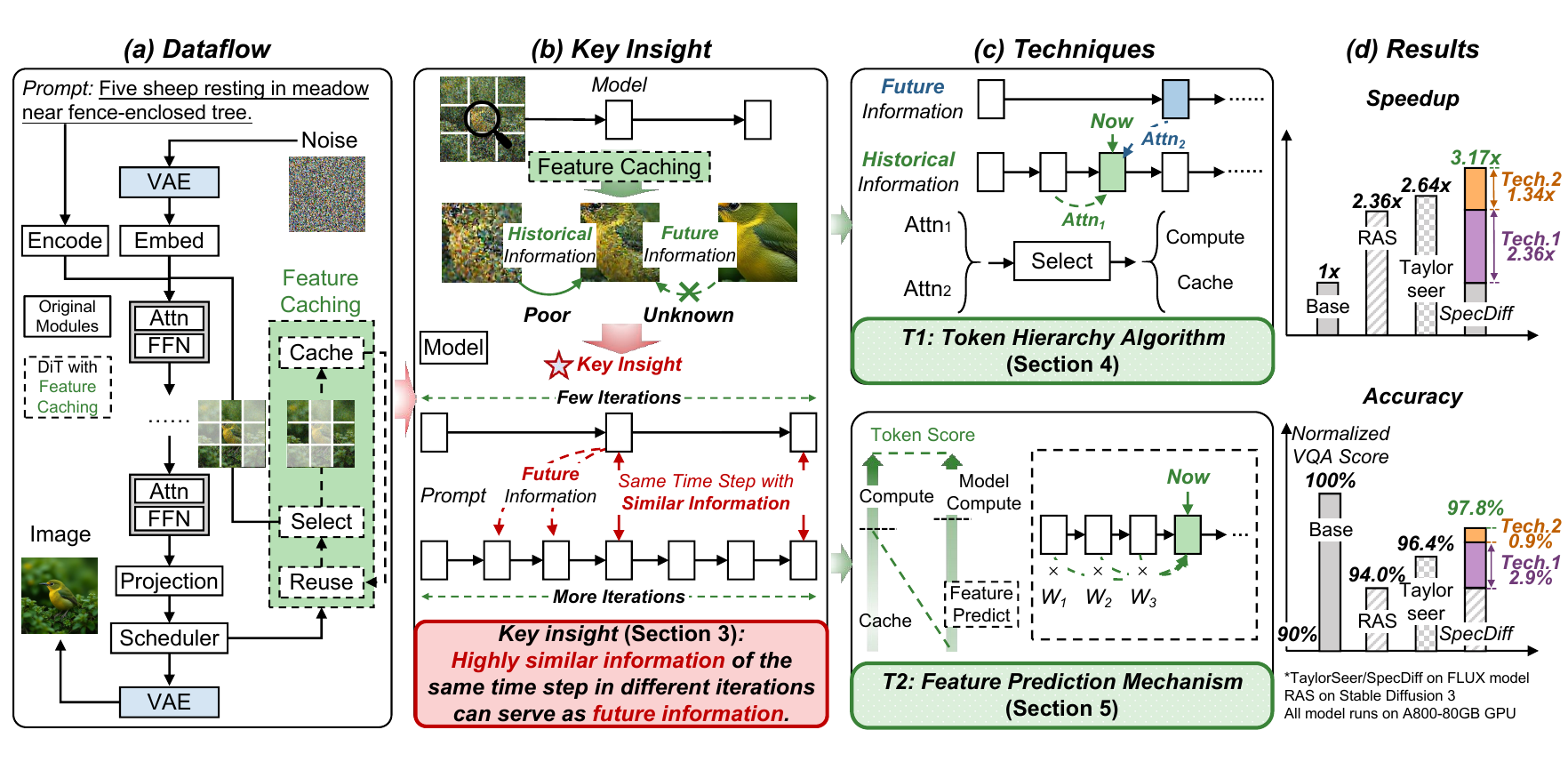} 
    \caption{Overview of \textit{SpecDiff}. (a) Dataflow of Diffusion Transformer with feature caching. (b) Key insight: Highly similar information from the same time step in different iterations can serve as future information. (c) Two main techniques of \textit{SpecDiff}. (d) Results on accuracy and speedup.}
    \label{fig:overview} 
\end{figure*}

Consequently, many previous works have explored techniques to reduce the computation workload, including algorithm optimization, system enhancements, and hardware advancements~\cite{ding2025vida,yuan2025vgdfr,zhang2025ditfastattnv2}. Recently, feature caching has become an emerging and promising method that exploits the similarities between features inherent in iterative inference of the diffusion model~\cite{ras,taylorseer}. By caching and reusing relatively invariant features to replace redundant model computations, it effectively reduces the overall computational workload and achieves significant acceleration. Furthermore, it is evident that the effectiveness of feature caching greatly depends on its design and implementation, specifically, on how to select the features that need to be cached precisely. This selection has a critical impact on both the accuracy and speedup of the diffusion model.

In this paper, we analyze feature caching \textbf{from a novel perspective, information utilization,} and point out that the information utilization is directly related to the design of the feature caching and significantly impacts the overall performance. Current work only focuses on exploiting historical information(\textit{e.g.}, feature similarity between historical steps~\cite{ras}) for feature caching and fails to predict the future characteristics of the model accurately. 
As shown in Figure~\ref{fig:overview}(b), historical information only captures the local variation and ignores potential image mutation, resulting in poor performance (over $60\%$ accuracy loss~\cite{selvaraju2024fora}) in some scenarios.


Existing work~\cite{gao2024mdtv2maskeddiffusiontransformer} has proven that the features focused by the diffusion model are strongly related to the time step parameter. This inspires the potential of models with future time steps to focus on information to assist in token selection. 
From Figure~\ref{fig:overview}(b), the
information of the same time step in different iterations is highly similar. Thus we propose \textbf{a novel paradigm using few self-speculation steps of the original model in advance to introduce future information} to capture the global variation.
To fully leverage speculative future information for precise feature caching, we present \textit{SpecDiff}, a training-free multi-level feature caching strategy based on speculative information, including a feature selection algorithm based on self-speculative information and a multi-level feature classification algorithm based on feature importance scores. The contributions are summarized as follows.

\begin{itemize}

    \item  \textbf{Feature selection algorithm based on self-speculative information}. As illustrated in Figure~\ref{fig:overview}(c)-T1, based on the self-speculative future attention information and historical attention information, \textit{SpecDiff} assigns dynamic importance scores to each token, and performs cached feature selection according to these importance scores.
    
    \item \textbf{Multi-level feature classification algorithm based on feature importance scores.}  \textit{SpecDiff} classifies tokens by leveraging differences of feature importance scores and introduces a multi-level feature calculation strategy. By employing distinct strategies for tokens at each level, potentially costly cumulative errors arising from simple reuse will be avoided.

    \item Extensive experiments show that \textit{SpecDiff}  achieves average $2.80\times$, $2.37\times$ and $3.17\times$ speedup with negligible quality loss on Stable Diffusion 3, 3.5~\cite{stablediff_3,stable_diff_3_5} and FLUX~\cite{FLUX} shown in Figure~\ref{fig:overview}(d). As shown in Figure~\ref{fig:pareto}, by merging speculative and historical information, \textit{SpecDiff} overcomes the speedup-accuracy trade-off bottleneck, \textbf{successfully pushing the Pareto frontier} in the efficient diffusion model inference.
\end{itemize}

\begin{figure*}[t] 
    \centering 
    \includegraphics[width=\textwidth]{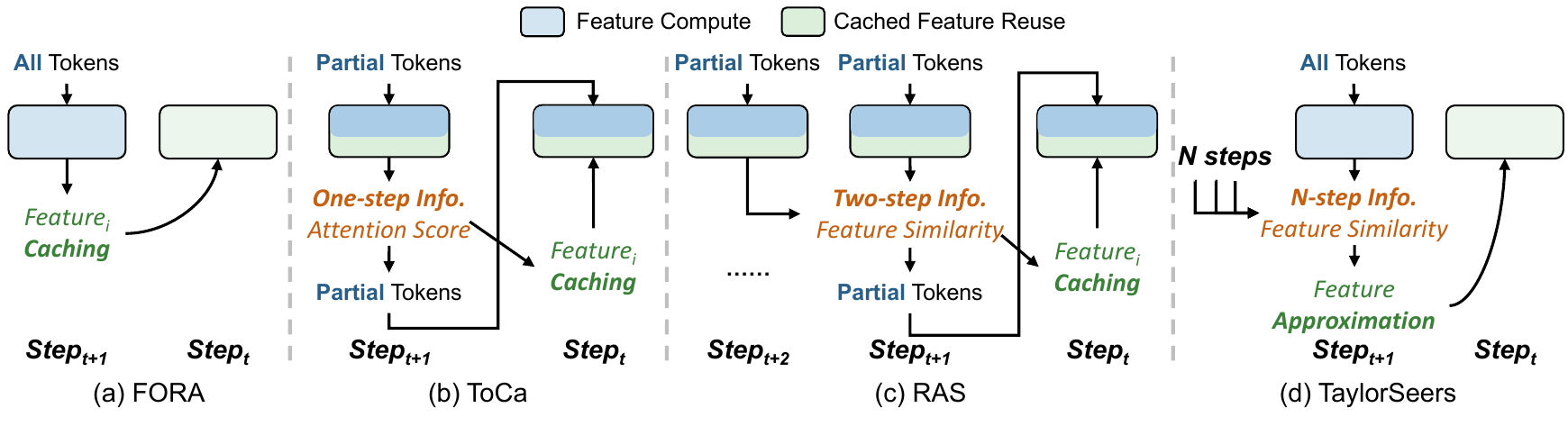} 
    \caption{Analysis on existing feature caching methods from the perspective of information utilization.}
    \label{fig:feature_cache}
\end{figure*}
\section{Background}
\subsection{Diffusion Model and Diffusion Transformer} \label{sec:back:diff}
As the typical neural network in multimodal artificial intelligence, the diffusion model has seen widespread application across diverse scenarios and has achieved impressive performance. Currently, diffusion transformer (DiT) architectures, represented by Stable Diffusion 3, 3.5~\cite{stablediff_3,stable_diff_3_5}, and FLUX~\cite{FLUX}, have become the mainstream design paradigm for diffusion models. Different from the traditional UNet-based diffusion models, 
the main composition of the DiT architecture is the transformer block with self-attention mechanism and feed-forward network(FFN) illustrated in Figure~\ref{fig:overview}(a). The self-attention mechanism elevates generation quality via long-range dependency modeling, while the FFN aims to capture inherent features, which compensate for the limitations of the self-attention mechanism.
Moreover, the variational auto encoder (VAE)~\cite{kingma2013auto} consists of an encoder and a decoder. The encoder is responsible for mapping the image from the pixel space to the latent space, while the decoder performs the reverse mapping. The text encoder and image embedding modules transform the text information and image information into high-dimensional features.


\subsection{Diffusion Transformer Acceleration}
In this paper, we focus on the DiT model inference acceleration. For the general matrix multiplication (GEMM) operator in FFN and self-attention operations of the DiT model, the GEMM ($M \times N \times K$) computation typically involves multiplying an input matrix ($M\times N$) with a weight matrix ($N\times K$). The ratio of computing workload and memory access (also known as arithmetic intensity) is defined as $\frac{M\times N\times K}{M\times K + N \times K}$. As illustrated in Figure~\ref{fig:overview}(a), the input of the DiT backbone in each iterative step is the matrix transformed by the noise image. The rows of the input matrix ($M$) commonly have comparable values with the rows of the weight ($N$) (\textit{e.g.}, $M=4096$ and $N=1536$ in Stable Diffusion 3~\cite{stablediff_3}), leading to the high arithmetic intensity. The arithmetic intensity reflects the computational efficiency, with higher values indicating that the GEMM is compute-bound rather than memory-bound.

Therefore, most of the works for DiT model acceleration mainly focus on reducing the computation. Model compression (\textit{e.g.}, distillation~\cite{yin2024one}, quantization~\cite{wu2025quantcache,wang2025maketrainingflexibledeploymentefficient}) exploits the inherent parameter features to make the model smaller, achieving computation reduction, but bringing expensive GPU training hours. Dynamic inference~\cite{zhang2023adadiff, xu2025specee} adaptively adjusts the number of model iterations based on the stability of intermediate results over iterations, leading to negligible quality loss but normal acceleration due to the overhead introduction. Feature caching methods~\cite{selvaraju2024fora,toca,ras,taylorseer} cache and reuse the features during iterations based on the similarity between token features, significantly reducing the computation and achieving high speedup with remarkable generation quality.

\subsection{Feature Caching} \label{sec:back:feature}
In this paper, we focus on the optimization of feature caching of the DiT model. Figure~\ref{fig:feature_cache} shows different designs of feature caching in existing mainstream works. FORA~\cite{selvaraju2024fora} directly caches and reuses the features in adjacent steps, resulting in profound quality loss in most scenarios (\textit{e.g.}, $>60\%$ accuracy drop in FID metric).
ToCa~\cite{toca} selects partial important tokens with high attention scores and caches the features of the left tokens in the current step for computation and reuse in the next step. 
RAS~\cite{ras} further proposes to utilize the feature similarity from two adjacent steps to achieve the important token selection. By identifying the importance of tokens, these two works have demonstrated notable results in both efficiency and accuracy.
Based on the differential similarity of all historical features, TaylorSeer~\cite{taylorseer} directly approximates the feature in the current step by using features from previous steps.

\begin{table}[!t]
\centering
\resizebox{0.45\textwidth}{!}{
\begin{tabular}{@{}cccc@{}}
\toprule
\textbf{Representative Works} & \textbf{Information Utilization} & \textbf{Speedup} & \textbf{Accuracy Loss} \\ 
\midrule
FORA & None & $1.8\times$ & $>60\%$ \\
ToCa & Historical One-step & $2.0\times$ & $\sim 30\%$ \\
RAS & Historical Two-step & $2.3\times$ & $\sim 20\%$ \\
TaylorSeer & Historical N-step & $2.6\times$ & $\sim 15\%$ \\
\bottomrule
\end{tabular}
}
\caption[Feature caching methods comparison]{Representative works on feature caching.}
\label{fig:feature_cache_comparison}
\end{table}

We analyze the underlying principles of these works above and point out that the differences in their feature caching designs can be attributed to different information utilization. Table~\ref{fig:feature_cache_comparison} presents the speed and accuracy of these designs from the perspective of information utilization, showing that both speed and accuracy improve as the amount of information used increases. Therefore, we conclude that the feature caching design is closely related to the information utilization, and that the overall speed and accuracy are directly proportional to the amount of information.

\begin{figure*}[!t] 
    \centering 
    \includegraphics[width=0.98\textwidth]{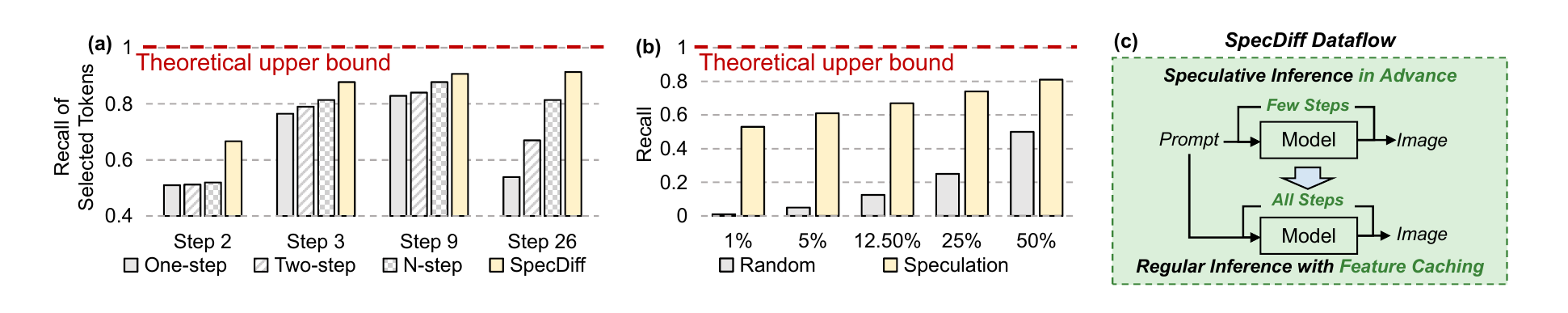}
    \caption{(a) The recall of selected tokens is defined as the proportion of tokens that are actually important in the time step being successfully predicted. When we add future information, we can get closer to achieving the theoretical upper limit of prediction. (b) Tokens that cannot be obtained by historical information will appear in future information. And the recall is much higher than the random method. (c) \textit{SpecDiff} uses a few-step inference to obtain future information for token selection.}
    \label{fig:insight_moti} 
\end{figure*}
To the best of our knowledge, \textit{SpecDiff}  is the first work to analyze feature caching through information utilization, revealing that the utilized information is closely related to the strategy design and critically affects the performance.

\section{Motivation}

\subsection{Key Challenges of Feature Caching}
The core of feature caching is to accurately compute the features of important tokens and reuse the features of unimportant tokens based on the available information. From the perspective of algorithm theory, the inference process of DiT essentially follows a Markov process, where the output in the next time step depends solely on the output in the current time step. As a result, existing works that rely exclusively on accumulating historical information remain confined within a subset of the total information available from previous steps, inherently limiting their potential performance gains. Experimentally, we evaluate the recall between the important tokens selected using historical information and the actual important tokens at each step, as shown in Figure~\ref{fig:insight_moti}(a). We find that there is still a significant gap from the upper bound regardless of one-step, two-step, or N-step information, revealing that historical information only captures the local variation shown in Figure~\ref{fig:overview}(b). Moreover, the recall does not improve significantly with the increase in the number of steps of historical information. On the contrary, due to the increase in the number of selected tokens, it slows down inference instead. 
Therefore, we argue that the key challenge of existing feature caching methods is \textbf{historical information is insufficient for accurately selecting important tokens}. And the key to the method is \textbf{how to introduce
future information to assist token selection effectively.}

\subsection{Key Insight}
Inspired by the speculation in LLM~\cite{li2024eagle,li2024specpim,xu2025specee}, we point out that speculation is an effective way to obtain future information by approximating the final result through limited model computation on the same input. 
Furthermore, as illustrated in Figure~\ref{fig:insight_moti}(a), our method assessing token significance through attention scores reveals significant differences in token importance at time step 2 between consecutive iterations. Historical information can only predict 50\% of important tokens. As shown in Figure~\ref{fig:insight_moti}(a), by analyzing high-attention tokens occurring at identical time steps across iterations versus adjacent time steps within the same iteration using a fixed selection proportion, we observe that important tokens cannot be predicted by historical information have a considerable proportion in important tokens predicted by speculative information compared with random prediction. This indicates that tokens deemed significant but unobtainable from historical information can instead be acquired through future information access.

From the perspective of the token selection in feature caching, performing a single denoising process with fewer iterations can serve as a potential strategy to expand the pool of candidate tokens, increasing the likelihood of identifying tokens that will become important in future time steps. In other words, if there is complete consistency between token importance and time step, we can precisely identify the critical tokens in each time step, thus achieving optimal performance.
Therefore, we propose \textbf{a novel paradigm that leverages few self-speculation steps of the original model in advance to introduce future information} to optimize feature caching to capture the global variation. This approach incurs minimal computational and memory overhead, requiring $<5\%$ additional inference time and $<0.1\%$ extra memory.
\begin{figure*}[htbp] 
    \centering 
    \includegraphics[width=1\textwidth]{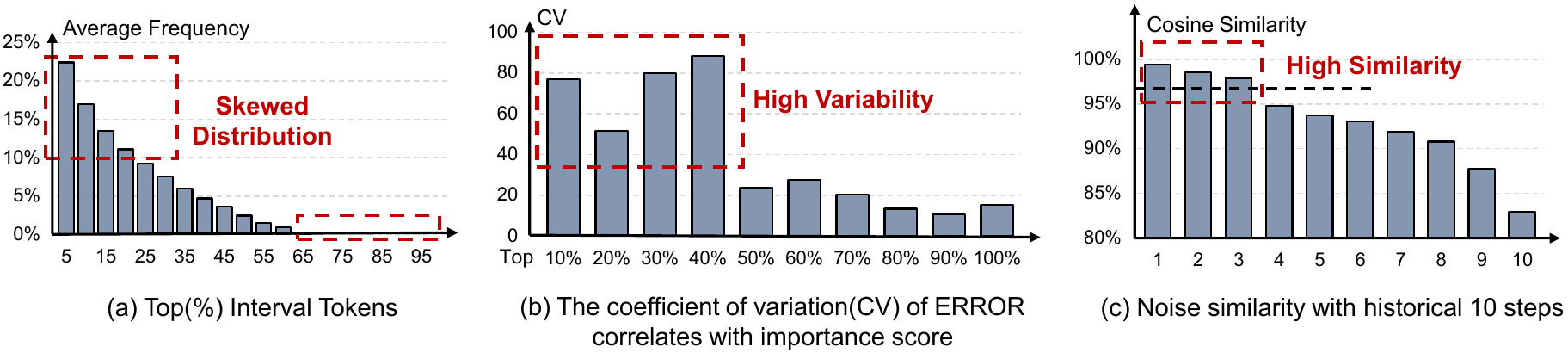} 
    \caption{(a) Without starvation scores, the distribution of tokens selected appears skewed. (b) Higher score tokens appear to have a higher ERROR coefficient of variation than lower ones. (c) Noise feature appears high similarity in continuous 3 steps. }
    \label{fig:method_process} 
\end{figure*}
\section{Methodology}
\label{sec:method}
\subsection{Feature Selection Algorithm}
As shown in Figure~\ref{fig:overview}(c)-T1, we determine token importance through attention scores, which is a widely recognized method. Building upon our previous exploration of historical information and self-speculative future information for token importance assessment, the key step lies in designing the calculation method for token importance scores.
We calculate the attention scores of the tokens across all layers in the previous iteration, and set the sum of these scores as the historical importance score of the token $x_i$ in the current iteration, denoted as $his(x_i)$. Similarly, we calculate the attention scores of future time steps obtained through self-speculative steps, set them as the future importance score of the token $x_i$ in the current iteration, denoted as $fut(x_i)$. Since the number of self-speculative steps is smaller than the number of full iterations, the time-step parameters cannot be perfectly aligned. For the calculation of the token importance score at the current time step of a full iteration, the token importance score of the nearest future time step is used as the future importance score. We define the token's importance score as the production of the historical score and the future score, given by the following formula:
\begin{equation}
    Score(x_i) = his(x_i) \cdot fut(x_i)
\end{equation}
For different cached token ratios $CR$ ($0<CR<1$)
, we select tokens with the largest importance scores in the quantity of $1-CR$
 as those that need to participate in network computation. For various caching ratios, we find that the number of times tokens are selected presents a significantly skewed distribution, as shown in Figure~\ref{fig:method_process}(a). For the case where only 20\% of important tokens are selected, the tokens with selection frequencies in the top 25\% account for more than 75\% of the selections. Moreover, this distribution exhibits a significant long-tail effect: approximately 40\% of tokens have never been considered important in dozens of iterations! Tokens that are cached too many times may have tremendous cumulative errors\cite{toca}. Therefore, to ensure that these tokens still have the possibility of being selected, we introduce a starvation score into the token's importance score, which is positively correlated with the number of times the token is cached, denoted as:
\begin{equation}
    star(x_i) = e^{ cf(x_i)}
\end{equation}
 where  $cf(x_i)$
 is the frequency that token $x_i$ has been cached.
 Thus, the complete token importance score is given by:
 \begin{equation}
     Score(x_i) = his(x_i) \cdot fut(x_i) \cdot star(x_i)
 \end{equation}

\subsection{Multi-level Feature Classification Algorithm}
After selecting a subset of tokens with the highest scores, how to efficiently approximate the features of the remaining tokens becomes a new key issue. Existing methods adopt the strategy of reusing the relevant strategy from the previous iteration\cite{toca,ras}, and the following simple equation can describe this process:
\begin{equation}
F(x_{t}^{\text{cached}}) \approx F(x_{t + 1}^{\text{cached}})    
\end{equation}  
and the relative error is:
\begin{equation}
\textit{ERROR} = \frac{F(x_{t}^{\text{cached}})}{F(x_{t + 1}^{\text{cached}})} - 1 
\end{equation}
Existing works rely on a strong assumption that the relative error $ERROR$ mentioned above approaches zero.
We examined the $ERROR$ for all tokens, as shown in Figure~\ref{fig:method_process}(b), which depicts the relationship between the top scores and $ERROR$. As shown in Figure~\ref{fig:method_process}(b), we found that the token's importance score exhibits a strong correlation with $ERROR$. Specifically, a higher token importance score corresponds to a larger variation coefficient for $ERROR$, which quantifies feature variability. Thus, directly reusing the features of these tokens would incur significant errors, leading to a decline in the final generation quality.
Moreover, for tokens with very low importance scores, the coefficient of variation of $ERROR$ appears lower, indicating that their features from the previous iteration can be directly reused.
Thus, under high caching rates (above 80\%), the feature reuse and approximation strategies of existing methods will lead to severe performance degradation, preventing them from maintaining good generation quality while achieving higher speedup effects. Here, we classify the token feature calculation strategies into the following three types:

\textbf{C1: Tokens that need to participate in network computation.} These tokens are those with the highest importance scores determined based on the caching rate.

\textbf{C2: Tokens that directly reuse features from the previous iteration.} We select tokens with the lowest scores, comprising 10\% of the total importance score, to adopt this strategy. For these tokens, directly reusing their features from the previous iteration can achieve a good approximation.

\textbf{C3: Tokens requiring approximate feature computation.} This class includes tokens that do not participate in network computation but fall within the top 90\% of the total importance score distribution. We examined the feature similarity of these tokens over time steps. As shown in Figure~\ref{fig:method_process}(c), the similarity between the noise at the current time step and the noise at the previous N time steps shows a downward trend. And due to the characteristics of high-dimensional space, when the cosine similarity is below 95\%, we consider that the noise no longer has a high similarity. Therefore, we only need to use the information from the previous three time steps to predict the noise at the current time step. And since the similarity of earlier time steps is lower, we adopt an approximation method of assigning different weights to the noise of different time steps. That is, for the earlier noise, we reduce its weight. Our approximation method can be expressed by the following formula:
$$
F(x_t^{\text{cached}}) = \sum_{i = 1}^{3} W_{t + i} \cdot F(x_{t + i}^{\text{cached}}), 
$$

\begin{equation}
\quad \text{where} \quad W_{t + i} = \frac{e^{-i}(T_{t + i}-T_t)}{\sum_{i = 1}^{3}e^{-i}(T_{t + i}-T_t)}    
\end{equation}
$T_{k}$ is the timestep of iteration $k$. $x_t^{\text{cached}}$ represents the network input corresponding to the tokens cached at time step $t$. $F(x_t^{\text{cached}})$ represents the cached noise corresponding to the cached token.

\section{Experiment}
\label{sec:exper}
\subsection{Experimental Configuration}
\textbf{Models and Datasets.} We evaluated the effectiveness of our method on the advanced DiT text-to-image models, Stable Diffusion 3~\cite{stablediff_3}, Stable Diffusion 3.5~\cite{stable_diff_3_5}, and FLUX.1 Dev~\cite{FLUX}. We randomly select 5000 text-image pairs from the COCO 2014 val dataset~\cite{lin2014microsoft} to generate 1024$\times$1024 images for the experiments. 

\textbf{Metrics and Baseline.} To evaluate the quality of the generated images and the alignment between text and images, we chose FID~\cite{DBLP:journals/corr/HeuselRUNKH17}, Clip Score~\cite{radford2021learningtransferablevisualmodels}, and VQA Score~\cite{lin2024evaluatingtexttovisualgenerationimagetotext} as our main experiment evaluation metrics. And for baseline comparison, we compared our method with the best similar method RAS~\cite{ras}, and the SOTA method of feature caching method TaylorSeer~\cite{taylorseer}. In order to comprehensively compare with the SOTA method TaylorSeer. We add SSIM, PSNR, and memory as new metrics. These baseline methods all have good generation quality with a good acceleration effect. We compared the generation quality of our method with these methods under different acceleration ratios. We set the parameters of baseline models Stable Diffusion 3~\cite{stablediff_3}, 3.5~\cite{stable_diff_3_5}, and FLUX~\cite{FLUX} for the default value. The inference step of our experiment is all set to 28. The CFG of Stable Diffusion 3 and 3.5 is set to 7.0, while the FLUX is 3.5.

\textbf{Hardware Platforms.} Our experiments were conducted on a machine with eight NVIDIA A800 80GB GPUs, and we conducted the generation speed experiment on a single A800 GPU. We implemented our method using the PyTorch framework and Diffusers libraries.

\begin{figure*}[!t] 
    \centering 
    \includegraphics[width=1\textwidth]{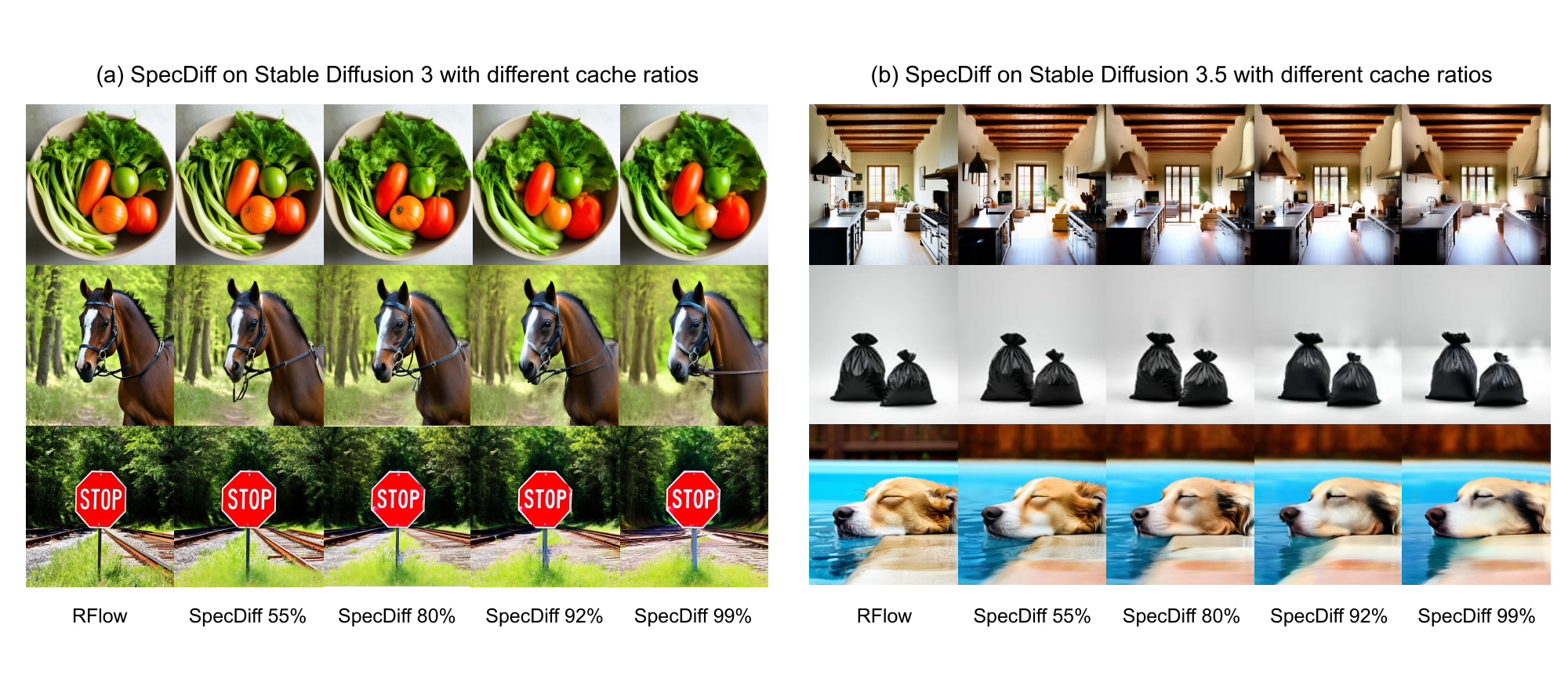} 
    \caption{Comparison between the images generated by \textit{SpecDiff} and the baseline images. \textit{SpecDiff} can maintain a considerable generation quality even with an ultra-high speedup ratio, especially highlighting the main objects in the images.}
    \label{fig:sd3_image} 
\end{figure*}

\begin{figure*}[!t] 
    \centering 
    \includegraphics[width=1\textwidth]{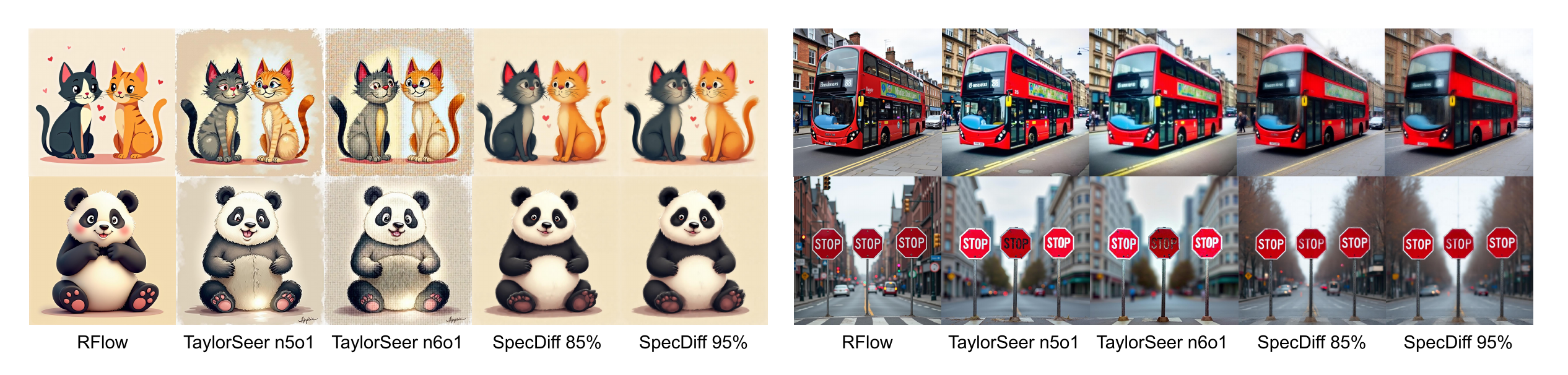} 
    \caption{Comparison between the images generated by \textit{SpecDiff} and those generated by Taylorseer on FLUX.1 Dev. \textit{SpecDiff} can better maintain consistency with the original image. Moreover, \textit{SpecDiff} can also maintain the alignment ability of the image text, and the key objects in the image have stronger consistency with the baseline.}
    \label{fig:flux_image} 
\end{figure*}

\subsection{Text-to-image Generation} \label{limit}
\textbf{Sampling efficiency improvement.} We evaluated the effectiveness of our method on advanced DiT models. As shown in Table~\ref{fig:sd3_exp_results}, \textit{SpecDiff} can maintain good generation quality while keeping a high speedup ratio. On Stable Diffusion 3, when the end-to-end speedup ratio is average 2.80 $\times$, the Clip Score only drops by 0.7\%, the FID only increases by 1.36\%, and the VQA Score drops by 3\%. On Stable Diffusion 3.5, when the end-to-end speedup ratio is average 2.37 $\times$, the Clip score only drops by 1.22\%, the FID only increases by 11.2\%, and the VQA score drops by 2.2\%. As shown in Table~\ref{fig:sd3_exp_results}, for various configurations of RAS, our method always significantly improves the quality while achieving a higher speedup. We evaluate \textit{SpecDiff} and TaylorSeer on FLUX using comprehensive metrics. As is shown in Table~\ref{fig:sd3_exp_results} and Table~\ref{fig:flux_comparison}, for various configurations of TaylorSeer, \textit{SpecDiff} outperforms TaylorSeer and achieves average $3.17\times$ speedup in terms of maintaining the alignment of image and text, human visual preference, image consistency with the original image, and memory usage. It is worth noting that the performance of our method on FID is not as remarkable as that on metrics such as Clip score. This may be because a relatively high CFG makes the style of the generated images tend to be unified, and ultimately, this unified style may not necessarily be consistent with real images. We selected and calculated these most important tokens, which strengthened this style characteristic and may lead to a relatively high FID value.

\begin{table}[t]
\centering
\resizebox{0.45\textwidth}{!}{
\begin{tabular}{c|ccccccc}
\toprule
\multicolumn{1}{c|}{\multirow{9}{*}{\rotatebox[origin=c]{90}{\textbf{Stable Diffusion 3}}}} & Method & Steps & Cached Ratio & FID$\downarrow$ & Clip Score$\uparrow$ & VQA Score$\uparrow$ & Speedup \\
\cmidrule{2-8}
& RFlow & 28 & 0 & 29.31 & 0.3176 & 0.9110 & 1.00$\times$\\
\cmidrule{2-8}
& RAS & 28 & 50\% & \textbf{27.26} & 0.3162 & 0.9005 & 1.61$\times$\\
& SpecDiff & 28 & 55\% & 27.57 & \textbf{0.3168} & \textbf{0.9057} & \textbf{1.61$\times$} \\
\cmidrule{2-8}
& RAS & 28 & 75\% & \textbf{27.38} & 0.3149 & 0.8849 & 2.09$\times$\\
& SpecDiff & 28 & 92\% & 29.52 & \textbf{0.3160} & \textbf{0.8888} & \textbf{2.43$\times$} \\
\cmidrule{2-8}
& RAS & 28 & 87.5\% & 40.92 & 0.3044 & 0.8611 & 2.40$\times$\\
& SpecDiff & 28 & 99\% & \textbf{29.75} & \textbf{0.3152} & \textbf{0.8822} & \textbf{2.80$\times$} \\
\midrule
\multicolumn{1}{c|}{\multirow{9}{*}{\rotatebox[origin=c]{90}{\textbf{Stable Diffusion 3.5}}}} & Method & Steps & Cached Ratio & FID$\downarrow$ & Clip Score$\uparrow$ & VQA Score$\uparrow$ & Speedup \\
\cmidrule{2-8}
& RFlow & 28 & 0 & 26.04 & 0.3190 & 0.9166 & 1.00$\times$ \\
\cmidrule{2-8}
& RAS & 28 & 50\% & 27.48 & 0.3167 & 0.9053 & 1.61$\times$ \\
& SpecDiff & 28 & 55\% & \textbf{27.17} & \textbf{0.3172} & \textbf{0.9105} & \textbf{1.62$\times$} \\
\cmidrule{2-8}
& RAS & 28 & 75\% & \textbf{28.12} & 0.3147 & 0.8873 & 2.13$\times$ \\
& SpecDiff & 28 & 92\% & 28.96 & \textbf{0.3159} & \textbf{0.8972} & \textbf{2.37$\times$} \\
\cmidrule{2-8}
& RAS & 28 & 87.5\% & 36.87 & 0.3132 & 0.8632 & 2.29$\times$ \\
& SpecDiff & 28 & 99\% & \textbf{30.08} & \textbf{0.3153} & \textbf{0.8891} & \textbf{2.74$\times$} \\
\bottomrule
\end{tabular}
}
\caption{Quality and speedup evaluation on SD 3 and 3.5}
\label{fig:sd3_exp_results} 
\end{table}
\begin{table}[h]
\centering
\resizebox{0.47\textwidth}{!}{ 
\begin{tabular}{@{}cclcccccc@{}}
\toprule
Method & Config & FID↓ & Clip Score↑ & VQA Score↑ & SSIM↑ & PSNR↑ & Memory(GB)↓ & Speedup \\ 
\midrule
RFlow & 0 & 27.68 & 0.3093 & 0.8986 & - & - & 38.36 & 1.00 $\times$\\
Taylorseer & N5O1 & \textbf{27.87} & 0.3090 & 0.8909 & 0.7098 & 16.93 & 42.66 & 2.47$\times$\\
SpecDiff & 85\% & 28.61 & \textbf{0.3125} & \textbf{0.8925} & \textbf{0.7101} & \textbf{19.20} & \textbf{41.46} & \textbf{2.52}$\times$ \\
Taylorseer & N6O1 & \textbf{28.83} & 0.3107 & 0.8822 & 0.6570 & 16.07 & 42.66 & 2.63$\times$ \\
SpecDiff & 95\% & 29.24 & \textbf{0.3124} & \textbf{0.8834} & \textbf{0.6963} & \textbf{19.02} & \textbf{41.46} & \textbf{3.17}$\times$ \\ 
\bottomrule
\end{tabular}
}
\caption{Comparison with TaylorSeer on FLUX.1 Dev }
\label{fig:flux_comparison}
\end{table}
\textbf{Design Space Exploration Experiment.}
We conduct experiments on the performance of \textit{SpecDiff} under different speculative steps. We conduct this experiment on Stable Diffusion 3. We set the number of iterations to 28 times each.
According to the data in Table~\ref{fig:supplementary_steps}, we can see that as the number of speculation steps increases, the quality of \textit{SpecDiff} generation will continuously improve, but what follows is a decrease in generation quality. When the speculation steps change from 2 to 4, for the case where the cached ratio is 99\%, the FID drops by 0.4\%, the Clip Score increases by 0.2\%, the VQA Score increases by 0.1\%, but the speed decreases by approximately 20\%. Therefore, a small number of speculation steps can achieve good results. Especially, when the speculation steps equal 2, \textit{SpecDiff} achieves the best trade-off between performance and speed.

\begin{table}[h]
\centering
\resizebox{0.45\textwidth}{!}{ 
\begin{tabular}{cccccc}
\toprule
Spec. Steps & Cached Ratio & FID $\downarrow$ & Clip Score $\uparrow$ & VQA Score $\uparrow$ & Speedup \\ 
\midrule
2 & 55\% & 27.57 & 0.3168 & 0.9057 & 1.61$\times$ \\
3 & 55\% & 27.52 & 0.3168 & 0.9059 & 1.45$\times$ \\
4 & 55\% & 27.43 & 0.3171 & 0.9066 & 1.33$\times$ \\ 
\midrule[0.5pt] 
2 & 92\% & 29.52 & 0.3160 & 0.8888 & 2.43$\times$ \\
3 & 92\% & 29.39 & 0.3161 & 0.8893 & 2.16$\times$ \\
4 & 92\% & 29.12 & 0.3163 & 0.8901 & 1.95$\times$ \\ 
\midrule[0.5pt]
2 & 99\% & 29.75 & 0.3152 & 0.8822 & 2.80$\times$ \\
3 & 99\% & 29.87 & 0.3154 & 0.8825 & 2.49$\times$ \\
4 & 99\% & 29.64 & 0.3157 & 0.8834 & 2.25$\times$ \\
\bottomrule
\end{tabular}
}
\caption{SpecDiff under different speculation steps on SD3}
\label{fig:supplementary_steps}
\end{table}
\textbf{Pushing the Pareto frontier of speed and quality.} As shown in Table~\ref{fig:sd3_exp_results} and Table~\ref{fig:flux_comparison}, our method is extensively evaluated against the RAS that supports the Stable Diffusion 3 series and the TaylorSeer that supports FLUX. Figure~\ref{fig:pareto} shows that our method outperforms them in terms of generation quality under the same computational load and end-to-end speedup. This indicates that our method can effectively capture the real tokens that the model focuses on during the image generation process, thus maintaining a more powerful generation quality while keeping a high end-to-end speedup ratio. It is worth noting that our method is particularly effective in maintaining the image-text alignment ability. This inspires us to improve the generation quality of the feature caching method by taking more global considerations of the diffusion denoising process, successfully pushing the Pareto frontier. Examples of images generated by \textit{SpecDiff} are detailed in Figure~\ref{fig:sd3_image} and Figure~\ref{fig:flux_image}. 
\begin{figure}[htbp] 
    \centering 
    \includegraphics[width=0.5\textwidth]{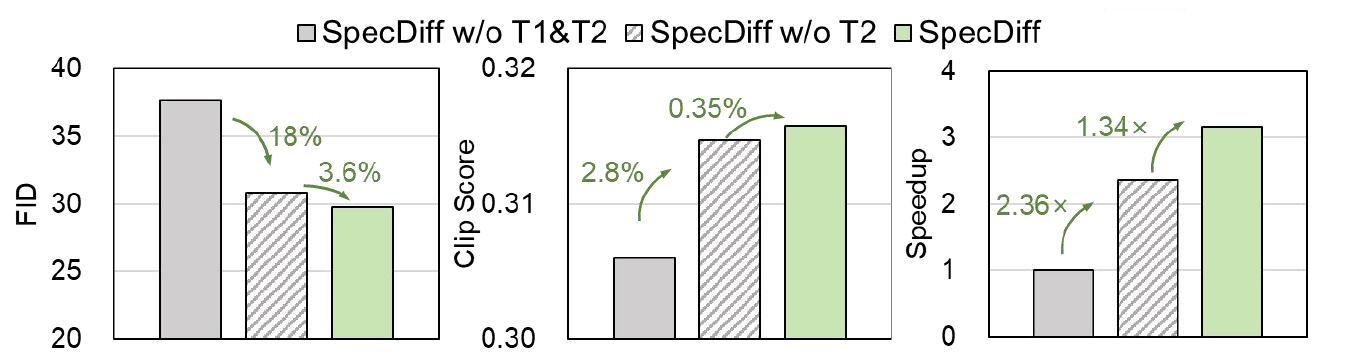} 
    \caption{Ablation study on speed and accuracy.}
    \label{fig:ablation} 
\end{figure}
\subsection{Ablation Study} \label{sec:ablation}

\textbf{Accuracy.} We conducted ablation experiments on the generation quality. The results in Figure \ref{fig:ablation} show that both the token hierarchy algorithm and the feature prediction mechanism can improve the generation quality of the model. Specifically, the token hierarchy algorithm reduces the FID by 18\% and increases the clip score by 2.8\%. The feature prediction mechanism further reduces the FID by 3.6\% and further increases the clip score by 0.35\%.

\textbf{Speedup.} We conducted ablation experiments on the generation speed. As shown in Figure \ref{fig:ablation}, both the token hierarchy algorithm and the feature prediction mechanism can improve the generation speed of the model. The feature prediction mechanism achieves an acceleration ratio of 2.36$\times$, and the token hierarchy algorithm further accelerates by 1.34$\times$. Eventually, an acceleration ratio of 3.17$\times$ is achieved.

\section{Conclusion}
\label{sec:conclusion}
In this paper, \textit{SpecDiff} analyzes the existing works on feature caching from a novel perspective, information utilization, and points out that current works only introduce historical information. Therefore, \textit{SpecDiff} proposes a novel paradigm using few self-speculation steps of the original model in advance to introduce future information. To fully leverage speculative future information, \textit{SpecDiff} proposes the feature selection algorithm and the multi-level feature classification algorithm. Extensive experiments show that SpecDiff achieves average 2.80$\times$, 2.74$\times$ and 3.17$\times$ speedup with negligible quality loss on Stable Diffusion 3, 3.5 and FLUX compared with RFlow on NVIDIA A800-80GB, successfully pushing the Pareto frontier.

\section{Acknowledgments}
This work was sponsored by Shanghai Rising-Star Program (No. 24QB2706200) and the National Natural Science Foundation of China (No. U21B2031).

\bibliography{aaai2026}

\end{document}